\documentclass[10pt,twocolumn,letterpaper]{article}

\usepackage[pagenumbers]{cvpr} 

\usepackage{graphicx}
\usepackage{amsmath}
\usepackage{amssymb}
\usepackage{booktabs}
\usepackage{adjustbox}
\usepackage{diagbox}
\usepackage{multirow}
\newcommand{\myparagraph}[1]{\vspace{3pt}\noindent{\bf #1}}
\usepackage[pagebackref,breaklinks,colorlinks]{hyperref}

\usepackage[symbol]{footmisc}
\usepackage[capitalize]{cleveref}
\crefname{section}{Sec.}{Secs.}
\Crefname{section}{Section}{Sections}
\Crefname{table}{Table}{Tables}
\crefname{table}{Tab.}{Tabs.}
\usepackage[table]{xcolor}

\DeclareUnicodeCharacter{2212}{-}
\makeatletter
\newcommand{\printfnsymbol}[1]{\textsuperscript{\@fnsymbol{#1}}}
\makeatother
\begin{document}

\title{Wiggling Weights to Improve the Robustness of Classifiers}
\author{Sadaf Gulshad\thanks{equal contribution} 
\quad Ivan Sosnovik \printfnsymbol{1}
\quad Arnold Smeulders \\
UvA-Bosch Delta Lab\\
University of Amsterdam, Netherlands\\
{\tt\small \{s.gulshad, i.sosnovik, a.w.m.smeulders\}@uva.nl}
}

\maketitle

\begin{abstract}

Robustness against unwanted perturbations is an important aspect of deploying neural network classifiers in the real world. Common natural perturbations include noise, saturation, occlusion, viewpoint changes, and blur deformations. All of them can be modelled by the newly proposed transform-augmented convolutional networks. While many approaches for robustness train the network by providing augmented data to the network, we aim to integrate perturbations in the network architecture to achieve improved and more general robustness. To demonstrate that wiggling the weights consistently improves classification, we choose a standard network and modify it to a transform-augmented network. On perturbed CIFAR-10 images, the modified network delivers a better performance than the original network. For the much smaller STL-10 dataset, in addition to delivering better general robustness, wiggling even improves the classification of unperturbed, clean images substantially. We conclude that wiggled transform-augmented networks acquire good robustness even for perturbations not seen during training. 
\end{abstract}

\section{Introduction}
\label{sec:intro}

\begin{figure}
    \centering
    \includegraphics[width=\linewidth]{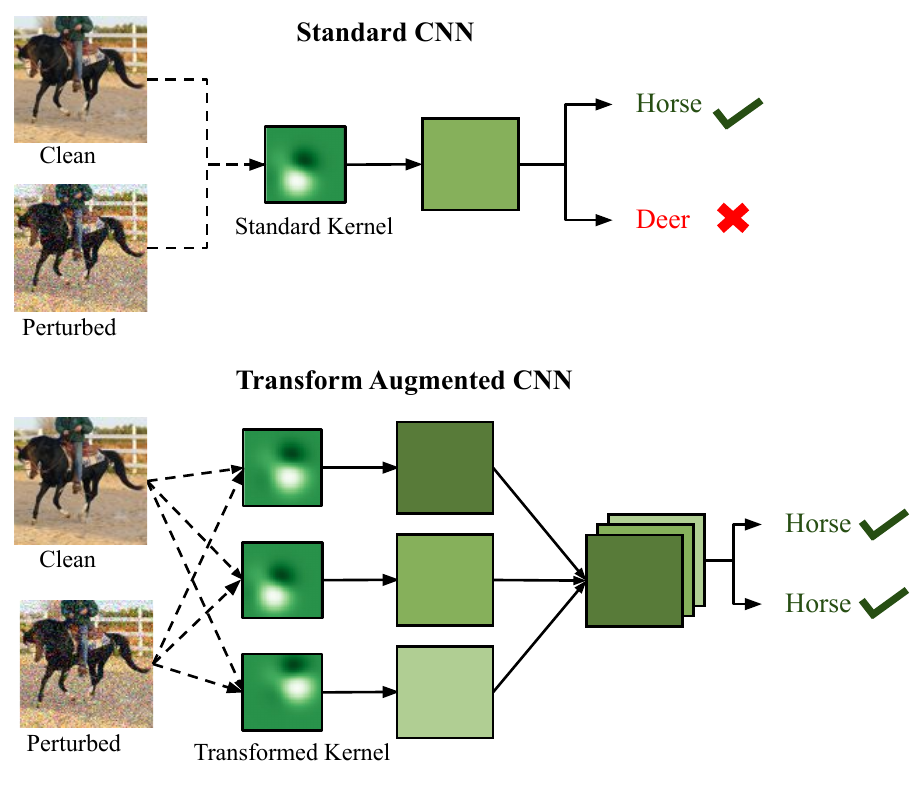}
    \caption{A standard neural network (top) and the transform augmented network (bottom) for classification. The transform augmented network has integrated perturbation transforms, which allow for better robustness against perturbations in the input without specialized training. }
    \label{fig:Teaser}
\end{figure}

Where most of the current network designs understandably focus on accuracy, we consider ways to robustify networks against unwanted perturbations, as robustness is important for building trust in the new technology. 

Neural network accuracy is affected by Gaussian noise or blur in the image, as is documented in \cite{azulay2018deep, dodge2017quality, recht2018cifar}. Also, occlusion or color saturation will have a profound but different effect on the network's performance. To achieve robustness, some train with perturbed rather than clean images \cite{schneider2020improving}. In \cite{rusak2020increasing}, training images are perturbed by an uncorrelated, learning noise generator. In \cite{robey2020model}, it was proposed to train the network with images from a generative model, while \cite{yin2019fourier} showed that training with noised perturbations help against high-frequency perturbations. Adversarial training \cite{goodfellow2014explaining} also belongs to the group of perturbed training methods. To date, it remains an open question whether adversarial training generalizes the robustness to a broad class of natural perturbations \cite{zhang2019limitations, engstrom2019exploring, gulshad2021natural}. In this work, we aim to contribute to general robustness without augmenting data. 

\begin{figure*}
    \centering
    \includegraphics[width=\linewidth]{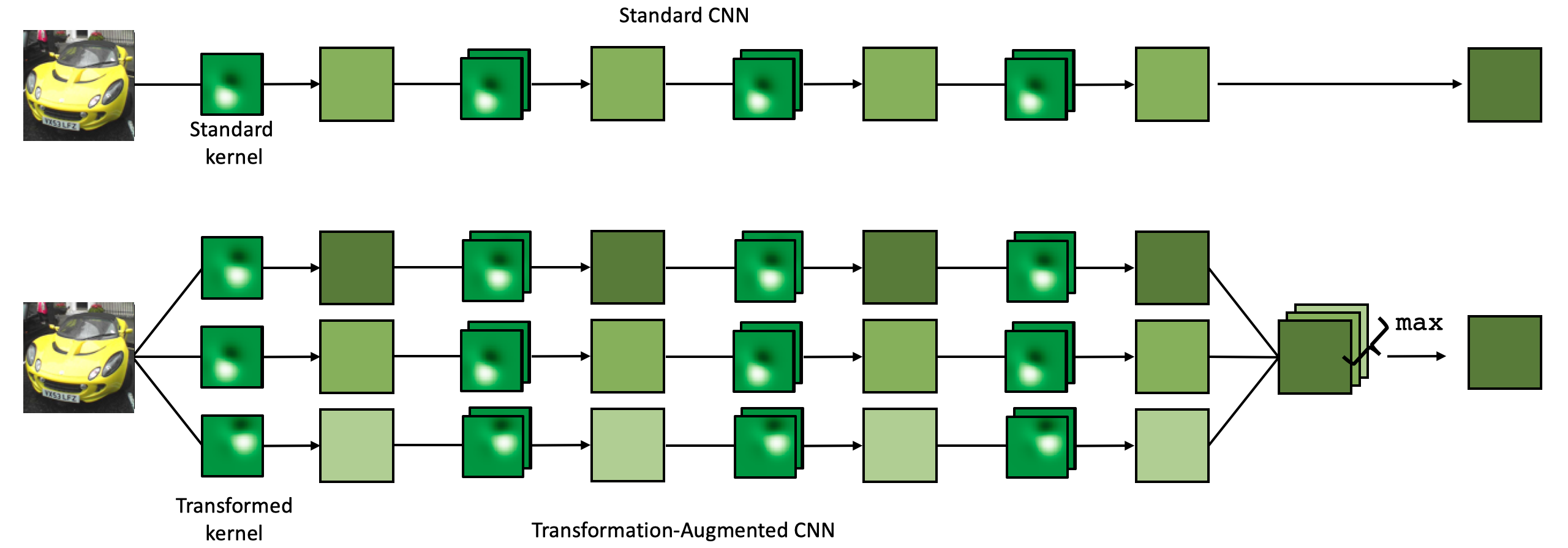}
    \caption{{Transform Augmented CNN Architecture. } Top: A standard CNN with four convolutional layers. Bottom: Its transform-augmented (TAConv) variant. By multiplying the fixed basis with the trainable weights, a single network is transformed in to a network with multiple paths, each path with a different basis. At the end, the maximum is selected. }
    \label{fig:Architecture}
\end{figure*}

Instead, we focus on modifying the network to enhance the generality of the robustness by considering the broad class of perturbations described by image filter transforms. As an example of our approach, we integrate six transformations in the network by wiggling the weights, see Figure \ref{fig:Architecture}. Apart from delivering better results on general robustness, the method also leaves future computational optimization and mathematical guarantees open.

We are inspired by \cite{jaderberg2015spatial,felzenszwalb2009object, dai2017deformable, cohen2016group}, building geometric transformations into the network to achieve equivariance. Effectively, equivariant networks deliver a solution for robustness against geometric transforms. We consider geometrical transformations to be robust against natural perturbations both global: {rotation-scaling perturbation}, {perturbation by object occlusion} and {Gaussian Blur} transformations, and local ones: {Gaussian Noise}, {Elastic deformations} and {Snow occlusion}. Together, they cover the breadth of 15 natural perturbations, as extensively documented in \cite{hendrycks2019benchmarking}. The six selections provide a good sample of realistic, real-world perturbations to which the network should be robust against, see Figure \ref{fig:sample_pert}. 

To permit a fair comparison on the effectiveness of network modifications for improved robustness, we tune the decay in the classification performance to be equal for all styles of perturbation. 
\begin{itemize}
    \item We propose Transform-augmented convolutions to integrate diverse perturbations for the enhancement of their general robustness.
    \item We demonstrate general robustness of our method on perturbations \textit{seen} during training and, more importantly, also on perturbations \textit{unseen} during training, including adversarial perturbations.
    \item Besides enhancing the general robustness against natural perturbations, we improve the classification performance on clean images, leading to the state-of-the-art on the STL-10 dataset, $95.45 \%$, and CIFAR-10, $94.97 \%$, without data augmentation.

\end{itemize}
 
\section{Related Work}
\label{sec:related_work}
\myparagraph{Vulnerability of Classifiers to Natural Perturbations.} In \cite{fawzi2015manitest,kanbak2018geometric} authors showed that neural networks are not even robust to translations and rotations. \cite{geirhos2017comparing} deduced that the performance of neural networks drops significantly as compared to humans with the increase of the signal-to-noise ratio of images. \cite{dodge2017study} also concluded that although neural networks are on par in performance with humans, they fail to perform well in the presence of perturbations like Gaussian noise or blur, which are easily handled by humans. Therefore, it is crucial to build robustness against such perturbations into the classification without degrading the performance on clean images, especially in such applications like autonomous driving and health. 

\myparagraph{Benchmarking Natural Perturbations.} To promote the study of robustness against naturally occurring perturbations, a few benchmarks have been proposed \cite{hendrycks2019benchmarking,hendrycks2020many,geirhos2017comparing}. Closely related to our work, in \cite{hendrycks2019benchmarking} the authors have introduced a large benchmark for natural perturbations, quite a few of which will be correlated \cite{laugros2021increasing}. In our work, we selected six more or less independent types of natural perturbations covering the breadth of styles, see Figure \ref{fig:sample_pert}. In the reference, the authors have gone through the effort of defining five levels of severity for each type of perturbation. These levels are based on the visual effect, but not standardized on their effect on the classification. As robustness is primarily aimed at the loss of classification performance, in this work at first we quantitatively standardize the comparison among differently trained networks to analyze the effect on their robustness. 
Table \ref{table:stand} shows the significance of our standardization method for fair comparison of robustness. When using the mean square error (MSE) between clean and perturbed images for standardization of perturbations, we see that the MSE shows a large variation in classification performance among different types of perturbations. Especially, the MSE calculated for adversarial and other natural perturbations show different behavior. This is because adversarial perturbations are generated in order to misclassify an image while keeping the optical difference between clean and adversarial images to a minimum. We start from  standardizing according to their effect on classification accuracy. Hence, we drop the accuracy of the network by a constant value for each type of perturbation, Table \ref{table:stand}. This enables the comparison among different perturbations and the robustness of classifiers. 
\begin{table}
\begin{center}
\begin{tabular}{l  c  c  }
\hline
Input & {Our standardization} & MSE   \\
\hline 
Adversarial  & 10.22 & 0.02 \\
Elastic    & 10.60 & 54.31 \\
Occlusion    & 10.24 & 199.73 \\
Gaussian Noise   & 10.10 & 11.79 \\
Wave    & 10.18 & 602.61  \\
Saturation    & 10.40 & 269.71  \\
Blur    & 10.51 & 18.20  \\
\hline
\end{tabular}
\end{center}
\caption{{Significance of our standardization for CIFAR-10.} To permit fair comparison, perturbation's parameters are tuned in such a way to standardize the drop in classification performance, here approximately $10\%$. Standardizing on the basis of the mean square error (MSE) between clean and perturbed images gives a large difference in classification performance, and hence is considered an unfair criterion for the purpose of robustness evaluation.}\label{table:stand}
\vspace{-0.5cm}
\end{table}

\myparagraph{Robustness to Natural Perturbations.} To improve the robustness against natural perturbations,  \cite{schneider2020improving} proposed to use batch normalization performed on perturbed images instead of clean ones. Similarly,  \cite{tang2021selfnorm} introduced two different normalization techniques, Selfnorm and Crossnorm to enhance the robustness against perturbations.  \cite{benz2021revisiting} also utilized perturbed samples and proposed to rectify batch normalization statistics for enhancing the robustness of neural networks against perturbations. 
Simultaneously,  \cite{rusak2020simple} introduced a noise generator that learns uncorrelated noise distributions. Training on these noisy images enhanced the performance against natural perturbations.  \cite{gulshad2021natural} trained on images with adversarial as well as natural perturbations like occlusions or elastic deformations, while achieving good generalization for many other unseen perturbations.  \cite{robey2020model} and  \cite{wong2020learning} argued that it is impossible to capture all possible natural perturbations mathematically. Therefore, they used generative models to generate images with perturbations to train the network. Instead of training with perturbed inputs, in this work we integrate predefined common perturbations into the network to enhance robustness.

\myparagraph{Robustness to Adversarial Perturbations.} In \cite{szegedy2013intriguing}, the authors explored the robustness of neural networks. They showed that by adding small amounts of carefully crafted noise i.e. \textit{adversarial perturbations} to the images it is possible to change the prediction of the classifier. Since then, plenty of research \cite{kurakin2016adversarial,papernot2016limitations,su2019one,carlini2017towards,moosavi2016deepfool} has been performed on finding different types of adversarial perturbations and study the robustification against them \cite{goodfellow2014explaining,kurakin2016adversarial,goodfellow2014explaining,carlini2017towards,dong2020benchmarking}. In this work, we utilize a strong yet undefended attack basic iterative method \cite{kurakin2016adversarial} for generating adversarial perturbations. \cite{rusak2020increasing} focused on robustification against adversarial as well as natural perturbations by using properly tuned Gaussian and Speckle noise. In this work, instead of generating tuned noise and then training the network, we build-in six different types of perturbations in the network which enhance the robustness against both natural and adversarial perturbations.

\myparagraph{Built-in Image Transforms.} 
The idea of analysing images together with their transformed versions has a long history. One of the first methods suggested modelling transformations as small units that locally transform their inputs for estimating geometric changes, i.e. capsules  \cite{hinton1981parallel}. In contrast, in \cite{simard1998transformation} neural networks are not modified and the update rule of the gradient descent is adapted to learn transformation-invariant weights. Later, both directions evolved a lot. In \cite{dai2017deformable,Worrall_2017_CVPR,cohen2016group,weiler2018learning,worrall2019deep,sosnovik2019scale,sosnovik2021disco,bruna2013invariant} the authors equip neural networks with an extra notion of either rotation or scale transformation for the purpose of better transformation processing in tasks where it is the main factor. In \cite{jaderberg2015spatial,laptev2016ti,kanazawa2014locally} neural networks are considered almost as block boxes and modifications are proposed to make them invariant under input transformations. 

While these methods considered specific geometric transformations to perform better analysis of downstream tasks, we focus on natural perturbations in general and demonstrate that any of them can be incorporated into a CNN for improved robustness.
\section{Method} 
\label{sec:method}
\subsection{Image Transformations}
Let us consider an image $f$. It can be reshaped as a vector $\mathbf{f}$. A wide range of image transformations can be parametrized by a linear operator: scaling, in-plane rotations, shearing. Other transformations, such as out-of-plane rotations, can not be parametrized in an image agnostic way. However, for small deviation from the original image Taylor expansions can be used, which gives a linear approximation for many image transformations of practical use. Indeed,
\begin{equation}
\label{eq:taylor}
\begin{split}
    T[f](\epsilon)
    & \approx T[f](0) + \epsilon\left.\Big(\frac{\partial T[f]}{\partial \epsilon}\Big)\right\rvert_{\epsilon=0} \\
   & = \mathbf{f}  + \epsilon\mathbf{L}_T\times \mathbf{f} 
   = (\mathbf{I} + \epsilon\mathbf{L}_T)\times \mathbf{f} 
   = \mathbf{T} \times \mathbf{f}
\end{split}
\end{equation}
where $T$ is a transformation, $\epsilon$ is the parameter of the transformation and $\mathbf{T}$ is a linear approximation of $T$ for small values of the parameter. For scaling the parameter is the logarithm of the scaling factor, for rotations it is the angle, and so on. $\mathbf{L}_T$ is a matrix representation of an infinitesimal generator of $T$. 
\begin{figure}
    \centering
    \includegraphics[width=\linewidth]{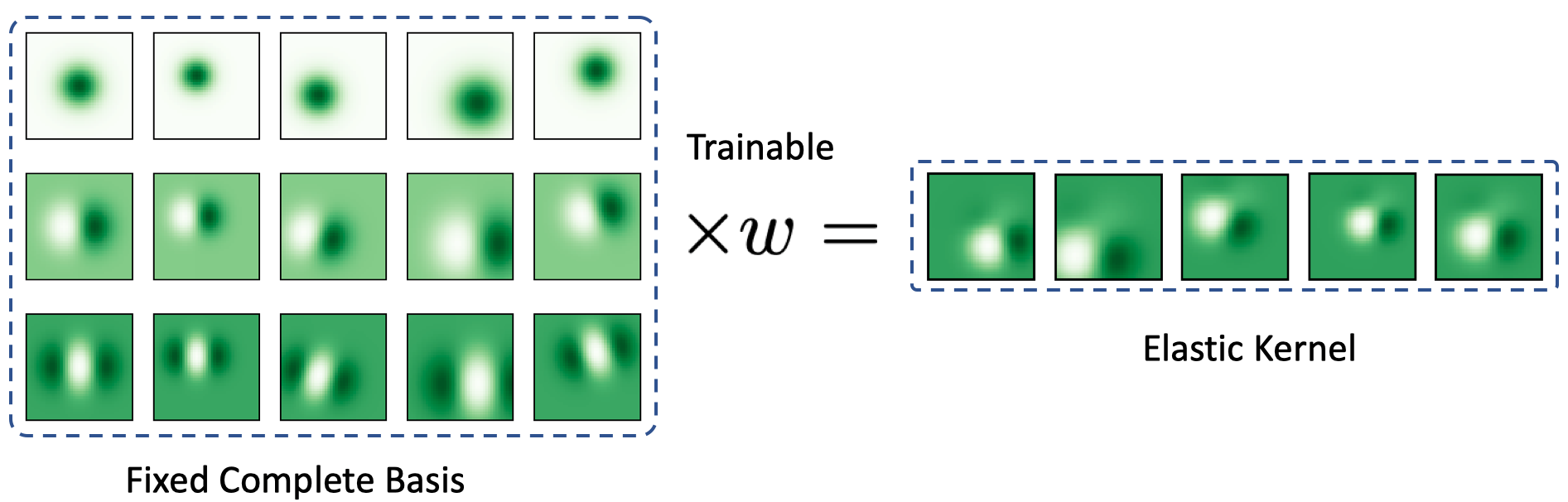}
    \caption{An illustration of how a set of transformed kernels is represented as a trainable linear combination of transform-augmented fixed basis functions.}
    \label{fig:basis}
\end{figure}
An image $f$ can also be viewed as a real-value function of its coordinates $f: x \rightarrow f(x)$. We focus here on transformations which can be represented by a smooth field of displacements $\tau$ in the space of coordinates. Equation \ref{eq:taylor} can then be rewritten as follows:
\begin{figure*}
    \centering
    \includegraphics[width=\linewidth]{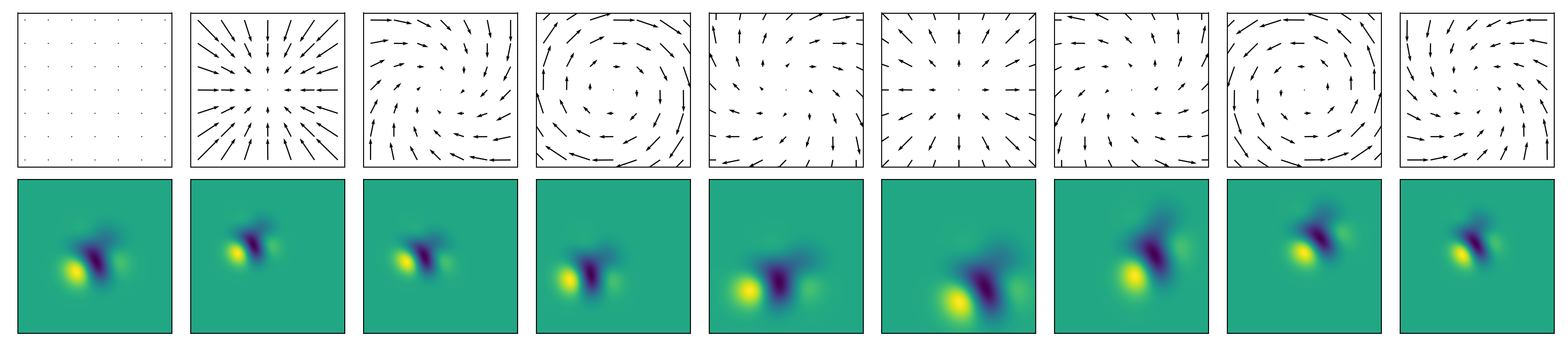}
    \includegraphics[width=\linewidth]{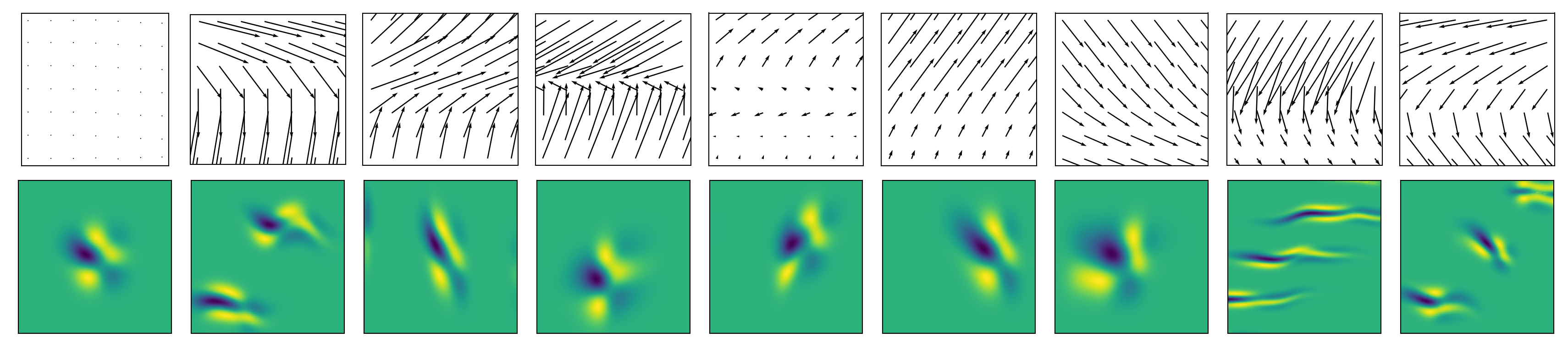}
\caption{Row top: smooth perturbations for the \textbf{global rotation-scaling transforms}. Row 2: an original filter and its transformed versions. Row 3: smooth perturbations for the \textbf{local elastic transforms}. Bottom row: an original filter and its transformed versions.}
      \label{fig:kernel_proj_global_local}
      \vspace{-3mm}
\end{figure*}
\begin{align}
\label{eq:smooth}T[f(x)](\epsilon) \approx f(x + \epsilon \tau(x))
\end{align}
We will refer to such transformations as elastic transformations. We will consider them as a linear approximation of a wide range of complex (camera) transformations. All other perturbations can be derived in a similar way up to an additive noise. We give a more detailed illustration of this derivation in supplementary materials.

\subsection{Transform-Augmented Convolutions}
Let us consider a convolutional layer $\Phi$ parameterized by a filter $\kappa$. It takes input image $f$. The output is:
\begin{equation}
    \label{eq:conv_output}
    \Phi(f, \kappa) = f\star \kappa = \mathbf{K} \times \mathbf{f}
\end{equation}
where $\mathbf{K}$ is a matrix representation of the filter. While, when data augmentation is used, a transformed version of the image can be fed as an input.
\begin{align}
\begin{split}
    \label{eq:conv_output_transformed}
    \Phi(T[f], \kappa) &= T[f]\star \kappa \\
    & = \mathbf{K} \times (\mathbf{T}\times\mathbf{f})  \\
   & = (\mathbf{K} \times \mathbf{T}) \times \mathbf{f} 
    = \Phi(f, T'[\kappa]) 
\end{split}
\end{align}
In the most general case, $\mathbf{K} \mathbf{T}$ is a matrix representation of a zero padding, followed by a convolution with a kernel and a cropping afterwards. The size of the kernel $T'[\kappa]$ depends on the nature of the transformation $T$. If the transformation is global, the kernel can be of a size bigger than the input image. We will consider only the cases when $T'[\kappa]$ is of the same or of a slightly bigger size than the original one.

To incorporate the data augmentation into the  convolutional layers of the network, we propose \textit{transform-augmented convolutions}, shortly TAConv, as follows:
\begin{equation}
    \text{TAConv} = \max
    \begin{bmatrix}
        \beta_0 \Phi(f, \kappa) \\
        \beta_1 \Phi(f, T_1[\kappa]) \\
        \vdots \\
        \beta_n \Phi(f, T_n[\kappa])
    \end{bmatrix}
\end{equation}
where $\beta_i$ are trainable coefficients. We initialize them such that $\beta_0=1$, and the rest are zeros. The maximum is calculated per pixel among different transformations of the kernel. At the beginning of training, the operation is thus identical to the original convolution with the same filter. If it is required during training, the other coefficients will activate the corresponding transformations.
\subsection{Transformations of a Complete Basis}
In order to apply transformations to filters, we parametrize each filter as a linear combination of basis functions:
\begin{equation}
    \label{eq:filter_basis}
    \kappa = \sum_i w_i \psi_i
\end{equation}
where $\psi_i$ are functions of a complete fixed basis and $w_i$ are trainable parameters. The approach is illustrated in Figure \ref{fig:basis}. We follow \cite{jacobsen2016structured} and choose a basis of 2-dimensional Gaussian derivatives. 

The transformations when applied to the basis form a transformed basis. Thus, for every transformation from the set, there is a corresponding transform basis. Weights $w_i$ are shared among all bases. We propose different sets of local and global transformations, we detail one global (rotations-scaling) and one local transformation (elastic) here, details for the rest of them i.e.  object occlusions, snow occlusions, Gaussian noise, and Gaussian blur are given in the appendix. 

Let us assume that the center of a filter is a point with coordinates $(0, 0)$. For every function from the basis, we first generate a grid of coordinates $(x, y)$. Then we evaluate the value of the function in the coordinates when projected on the pixel grid. 

\myparagraph{Global Rotation-scaling.} In order to transform the functions, we add a small displacement to the coordinates, which leaves the center untransformed. Given a grid of coordinates $(x,y)$, $\alpha$ the deformation intensity and $\sigma$ be the scaling factor, we define rotation-scaling (See Figure \ref{fig:kernel_proj_global_local}, Row top, 2) displacements as follows:
\begin{align}
\label{eq:rot_scale}
    x'&= x+\alpha(x\text{cos}(\theta)+y\text{sin}(\theta)) \\
    y'&= y+\alpha(-x\text{sin}(\theta)+y\text{cos}(\theta))
\end{align}
where $x',y'$ are the displaced coordinates. And $\theta$ is the scale-rotation parameter. When $\cos(\theta)$ is equal to 0 the whole transformation parametrizes rotation. When $\sin(\theta)$ is equal to 0 then it performs scaling. For all other cases, the transformation is a combination of both. The elasticity coefficient controls the severity of the transformations. Thus, for the case of rotation, it is a linear approximation of the $\sin$ of the rotation angle. For the case of scaling, $\alpha$ the scaling coefficient.

\myparagraph{Local Elastic Transform.}  Given a grid of coordinates $(x,y)$, $\alpha$ the elasticity coefficient and $\sigma$ be the scaling factor we define the elastic transformed filter as following (See Figure \ref{fig:kernel_proj_global_local}, Row 3, bottom), i) we take a 2D affine transform $A_\theta$ and map the coordinates $(x,y)$ to the target coordinates $(x^t,y^t)$:
\begin{align}
\label{eq:elastic}
    A_\theta \begin{pmatrix} x^t\\ y^t\\ 1 \end{pmatrix} = \begin{bmatrix}
        \theta_{11} \,\, \theta_{12} \,\, \theta_{13} \\
        \theta_{21} \,\, \theta_{22}\,\,  \theta_{23} \\
        \theta_{31}\,\, \theta_{23}\,\,  \theta_{33}
    \end{bmatrix} \begin{pmatrix} x^t\\ y^t\\ 1 \end{pmatrix}
\end{align}
In order to find $\theta$ parameters we select three points in the input grid $(x,y)$ and map them to the output $(x^t=x+U(-\alpha,\alpha), y^t=y+U(-\alpha,\alpha))$. Where, $U$ is the uniform distribution.
ii) We get another set of displaced coordinates $(x',y')$ by mapping the coordinates of the kernel as follows:    
\begin{align}
\label{eq:elastic}
    x' &= x+\alpha(\frac{1}{\sqrt{2\pi \sigma^2}}e^{-\frac{x^2}{2\sigma^2}}) \\
    y' &= y+\alpha(\frac{1}{\sqrt{2\pi \sigma^2}}e^{-\frac{y^2}{2\sigma^2}})
\end{align}
iii) Finally, we map the target coordinates $(x^t,y^t)$ to $(x',y')$ using bilinear interpolation. 

We follow \cite{sosnovik2019scale} and use a basis of 2 dimensional Hermite polynomials with the Gaussian envelope for all transforms:
\begin{equation}
\label{eq:herm}
    \psi_\sigma(x',y')= A \frac{1}{\sigma^2}H_n\left(\frac{x'}{\sigma}\right)H_m\left(\frac{y'}{\sigma}\right)\text{exp}\left[-\frac{x'^2+y'^2}{2\sigma^2}\right]
\end{equation}
where, $A$ is the normalization constant, $H_n$ is the Hermite polynomial of $n-$th order and $\sigma$ is the scaling factor. We iterate over $n,m$-pairs to generate functions. 
\begin{figure*}
    \centering
    \includegraphics[width=\linewidth]{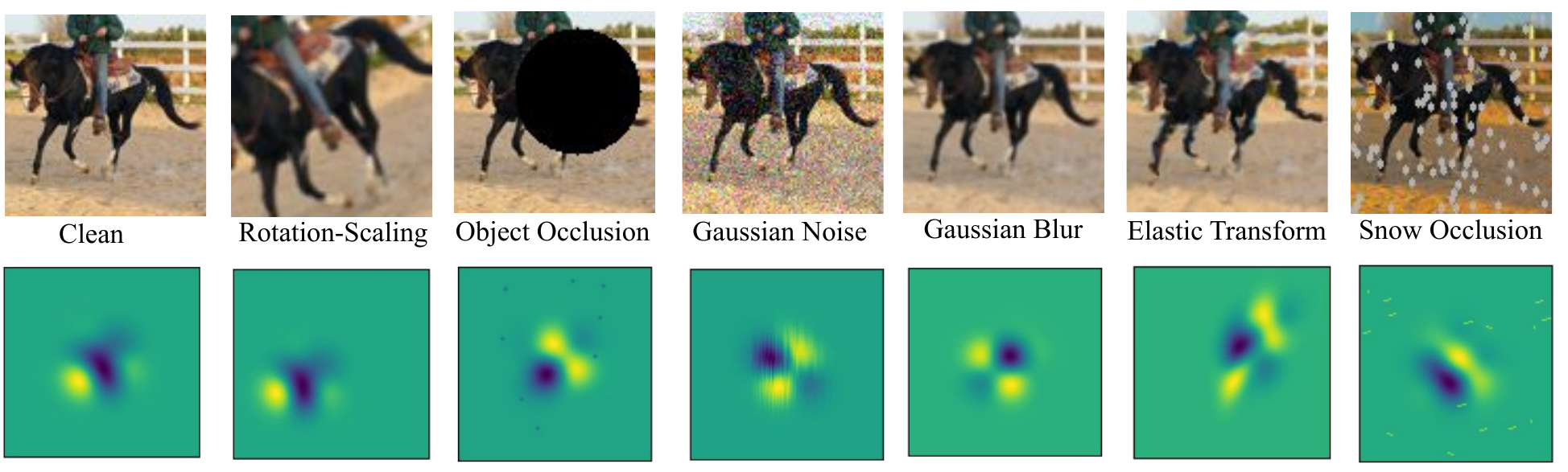}
    \caption{Top: Sample image from STL-10 dataset showing clean and six different natural perturbations. Bottom: Their respective filter representations in the network.}
    \label{fig:sample_pert}
\end{figure*}
\begin{table*} 
\begin{center}
\resizebox{\linewidth}{!}{\begin{tabular}{l  c  c  c c c  c  c }
\hline
  & &\multicolumn{6}{c}{\textbf{Transform Augmented Convolutions for clean inputs}}   \\ \cline{3-8}
\backslashbox{Test Input}{Network}  & \textbf{Standard Net} &  \textbf{Rotation-scaling}   &  \textbf{Object Occlusion}      & \textbf{Gaussian Blur}    &  \textbf{Gaussian Noise} & \textbf{Elastic} & \textbf{Snow Occlusion}  \\\hline 
Clean (CIFAR-10) & \textcolor{blue}{92.53} & {{94.50}}  & {94.37}  & {94.13} & {94.30} & {\textbf{94.97}} &{93.96}\\ 
Clean (STL-10)    & \textcolor{blue}{84.40}  & {94.12}            & {{95.29}} & {94.67} & {94.33} & {\textbf{95.45}} &{94.49}\\\hline

\end{tabular}}
\end{center}
\caption{Transform augmented convolutional networks achieve an improvement in the performance on clean data for CIFAR-10, and a significant improvement of $\approx 11\%$ for clean STL-10 data. This shows that our TAConv enhances the generalization on clean images. }\label{table:clean1}
\end{table*}
\subsection{Transform-Augmented Residual blocks}
In order to transform residual networks, we propose a straightforward generalization of the proposed convolution. The standard residual block can be formulated as follows:
\begin{equation}
    \text{ResBlock} =  f + G(f, \kappa_1, \kappa_2, \dots)
\end{equation}
The according augmented block is formulated as follows:
\begin{equation}\label{eq:EAResBlock}
    \text{TAResBlock}=  f + \max
    \begin{bmatrix}
        \beta_0 G(f, \kappa_1, \kappa_2, \dots) \\
        \beta_1 G(f, T_1[\kappa_1], T_1[\kappa_2], \dots) \\
        \vdots \\
        \beta_n G(f, T_n[\kappa_1], T_n[\kappa_2], \dots)
    \end{bmatrix}
\end{equation}
Transformed kernels augmented in the network architecture are shown in the Figure.\ref{fig:Architecture}. 

\subsection{Weights Transfer.}  In order to train neural networks successfully, initializing neural networks with Imagenet pretrained model weights is a common practice. However, it is not straight forward to transfer the weights of a standard network to our TAConv network because our network is composed of fixed basis and trainable weights, i.e. multiple parallel networks connected to each other Figure \ref{fig:Architecture}. Inspired by  \cite{sosnovik2021scale} we assume that in TAConv there is a subnetwork which is identical to the standard network, hence, we can transfer the weights of the standard network to our TAConv subnetwork. 
We start by disconnecting parallel networks by initializing all the weights responsible for inter correlations to zero. Now, the TAConv network until the TAConv max pooling layer (equation \ref{eq:EAResBlock}) consists of several parallel networks disconnected to each other. Convolutional layers of TAConv for which filter sizes match with the standard network, we initialize them with the weights from the standard network. $1\times1$ convolutions of the standard network and the TAConv network are identical, therefore, we copy the weights from the standard to the TAConv network. 
\vspace{-3mm}
\section{Experiments and Results}
\label{sec:exp}
\myparagraph{Datasets.}
Two datasets of varying input image sizes i.e. CIFAR-10 $32\times32$ pixels, STL-10 $96\times96$ pixels are considered in our experiments. CIFAR-10 consists of ten classes with 50000 training and 10000 test images \cite{krizhevsky2009learning}. STL-10 contains 5000 training and 8000 test images in ten categories \cite{coates2011analysis}.  

\myparagraph{Evaluating the Standard Network} We begin by training standard networks for each dataset on clean images. We fine-tune Resnet-152 networks pretrained on Imagenet and achieve $92.53\%$ and $84.40\%$ for CIFAR-10 and STL-10 clean test set respectively, see Table \ref{table:clean1}. We use cyclic learning rate scheduler with the learning rate of $0.05$. The only data augmentation used while training is random horizontal flip.  

\myparagraph{Standardizing Network Robustness.}
While considering standard networks as the baseline, we standardize the comparison among robustness of different networks by setting the desired drop at $10\%$ for each dataset, shown in Table \ref{table:pert1}.  We succeed in reaching a standardized drop with a maximum standard deviation of $0.44$. Hence, our standardization enables fair comparison among robustified networks on different types of perturbations.

\subsection{ Evaluating Transform Augmented Convolutional Network} 
We train each classifier network with transform augmented convolutions. For both datasets, we initialize the weights of TAConv Resnet-152 with Imagenet weights and fine-tune it. We add transform augmented convolution to the first convolutional layer with four different versions of a transform throughout the network as shown in Figure \ref{fig:Architecture}. 

\begin{table*} 
\begin{center}
\resizebox{\linewidth}{!}{\begin{tabular}{l  c  c  c c c  c  c  }
\hline
  & &\multicolumn{6}{c}{\textbf{Transform Augmented Convolutions for seen (on-diagonal) and unseen (off-diagonal) naturally perturbed images}}   \\ \cline{3-8}
\backslashbox{Test Input}{Network}  & \textbf{Standard Net} &  \textbf{Rotation-scaling}   &  \textbf{Object Occlusion}      & \textbf{Gaussian Blur}    &  \textbf{Gaussian Noise} & \textbf{Elastic} & \textbf{Snow Occlusion}  \\\hline 
 & &\multicolumn{5}{c}{\textbf{{CIFAR-10}}}  \\ \hline
 Rotation-scaling& \textcolor{blue}{82.81}  & 82.73 &   82.56 & {84.63} & {\textbf{87.00}} & {86.59} & { 85.97} \\

Object Occlusion     & \textcolor{blue}{81.90}  & 78.35  & 77.48  & 80.31 &  76.95          & 80.18                      & {81.81} \\
Gaussian Blur  & \textcolor{blue}{82.60}  & {{87.96}} & {85.56}  & {87.44} & {88.42} & {\textbf{90.03} } & {87.72}  \\ 
Gaussian Noise    & \textcolor{blue}{81.47}  & {88.96}  & {87.51} & {\textbf{90.64}} & {{89.72}} & {{89.38}} & {83.67} \\

Elastic      & \textcolor{blue}{82.61}  & {{85.62}} &  {82.70}  & {84.01} & {85.91} & {{87.12}} & {\textbf{86.09}} \\
Snow Occlusion  & \textcolor{blue}{82.81} & {83.86}   &     {83.12 }           & {{84.99}} & {\textbf{85.63}} & {{85.23} } & {{85.82}} \\ \hline
 & &\multicolumn{5}{c}{\textbf{{STL-10}}}  \\ \hline
  Rotation-scaling & \textcolor{blue}{73.98} & {89.48} & {\textbf{92.91}} & {91.77} & {{91.27}} & {92.29} & { 91.62} \\

Object Occlusion    & \textcolor{blue}{73.28}  &  {74.56}  &  {{85.58}}  &  {\textbf{84.04}} & {77.09}  & {{81.90}} & {{79.27}} \\ 
Gaussian Blur  & \textcolor{blue}{73.86}  &  {87.41}  &  {{90.85}}  &  {90.70} & {90.33}  & {\textbf{91.01}} & {89.76}  \\
Gaussian Noise    & \textcolor{blue}{73.60}  &  {90.70}  &  {86.91}  &  {{92.03}} & {89.65}  & {\textbf{92.60}} & {90.18}  \\

Elastic       & \textcolor{blue}{73.88}  &  {87.96}  &  {89.30}  &  {{89.91}} & {89.92}  &  {{90.94}} & {\textbf{89.99}}  \\
Snow Occlusion  &\textcolor{blue}{73.49}  &  {90.49}&           { \textbf{91.96 }  }         & {86.88} & {89.58}  & {85.63} & {{90.15} } \\
\hline
\end{tabular}}
\end{center}
\caption{{Evaluating Transform augmented convolutions for seen (on-diagonal) and unseen (off-diagonal) naturally perturbed images.} For a standard network, we drop the performance to, $\approx 10\%$ by adding perturbations. Our TAConv recovers the drop in the performance for all the perturbations on CIFAR-10 except Occlusion. For STL-10 besides all other perturbations we also recover the drop due to occlusions. This shows that our TAConv enhances the robustness against natural perturbations. }\label{table:pert1}
\end{table*}

\begin{table*} 
\begin{center}
\resizebox{\linewidth}{!}{\begin{tabular}{l  c  c  c c c  c  c  }
\hline
  & &\multicolumn{6}{c}{\textbf{Transform Augmented Convolutions for adversarially perturbed images}}   \\ \cline{3-8}
\backslashbox{Test Input}{Network}  & \textbf{Standard Net} &  \textbf{Rotation-scaling}  &  \textbf{Object Occlusion}      & \textbf{Gaussian Blur}    &  \textbf{Gaussian Noise} & \textbf{Elastic} & \textbf{Snow Occlusion}  \\\hline 
 & &\multicolumn{5}{c}{\textbf{{CIFAR-10}}}  \\ \hline
Adversarial                & \textcolor{blue}{82.13}  & {86.42}  & {83.40}  & {\textbf{87.58}} &   {86.23} & {{86.49}} &  {83.90}\\
Adversarial ($\epsilon=2$)   & \textcolor{blue}{28.48}  & 28.27  & 25.23  & {\textbf{37.15}} & {{28.57}} & {{28.48}} & {24.73}\\
\hline
 & &\multicolumn{5}{c}{\textbf{{STL-10}}}  \\ \hline
Adversarial                & \textcolor{blue}{71.91}   & {88.12}  &  {{90.62}} & {{90.28}} &  {88.29}& {\textbf{91.72}} & {89.68}\\
Adversarial ($\epsilon=2$)   & \textcolor{blue}{27.04}  & {{40.35}} & {35.21}  & {{40.29}} & {40.79}& {\textbf{50.06}} & {46.04}\\
\hline
\end{tabular}}
\end{center}

\caption{{Evaluating Transform augmented convolutions for adversarially perturbed images.} As before, for a standard network, we drop the performance by adding adversarial perturbations to cause, $\approx 10\%$ drop and a higher drop with $\epsilon=2$. For CIFAR-10, our Gaussian blur TAConv recovers the drop in the performance for both drop levels. For STL-10, Elastic TAConv recover best at $\approx 23\%$. TAConv enhances the robustness against adversarial perturbations. }\label{table:adv1}
\vspace{-3mm}
\end{table*} 

\subsubsection{Evaluating Robustified Networks on Clean Images.} For both datasets, we experiment by augmenting the first convolutional layer with rotation-scaling, object occlusion, Gaussian blur, Gaussian noise, elastic and snow occlusion convolutions one-at-a-time.  Introducing TAConv only to the first convolutional layer showed an improvement in the performance. For clean CIFAR-10 test set, elastic augmented convolutions showed the best performance with an improvement of $2.44\%$, rotation-scaling following it with an improvement of $1.94\%$, see Table \ref{table:clean1}.

For STL-10, the improvement in the performance is significant, with elastic augmented convolutions, leading to an improvement of $11.05\%$ and object occlusion being the second best with an improvement of $10.89\%$, see Table \ref{table:clean1}. We contend that the reason behind the significant improvement in the performance for STL-10 dataset is that, STL-10 is a small dataset, and our augmented convolutions provide variations in the network similar to data augmentation, which leads to an improvement in the performance especially for small datasets. 
 
\subsubsection{Evaluating Robustified Networks on Seen Naturally Perturbed Images.}

Table \ref{table:pert1} (on-diagonal) compares the performance of a standard and our transform augmented networks in the presence of seen naturally perturbed images. Sample test perturbations are shown in Figure \ref{fig:sample_pert}. 

We observe that for CIFAR-10, our transform augmented convolutions recover the drop in the performance on seen perturbations except rotation-scaling and object occlusions. Gaussian noise augmented convolutions show the best recovery in the drop when tested on Gaussian. 

On the other hand, on STL-10, our transform augmented convolutions recover the drop for all the seen perturbations, with elastic augmented convolutions showing the best performance when tested on elastic perturbations.  Hence, our proposed transform augmented convolutions are generally robust against seen natural perturbations. 

\subsubsection{Evaluating Robustified Networks on Unseen Naturally Perturbed Images.}
On the CIFAR-10, we observe that all the transform augmented networks recover the drop in the performance for all the perturbed unseen inputs except occlusions, see Table \ref{table:pert1}. For Gaussian noise perturbations, Gaussian blur augmented convolutions, for rotation-scaling and snow perturbations  Gaussian noise augmented convolutions, and for Gaussian blur elastic augmented convolutions, and for elastic perturbations snow augmented convolutions showed the best recovery in the drop of the performance. Thus, our proposed transform augmented convolutions show robustness against unseen natural perturbations except occlusions for CIFAR-10 dataset. We argue that the lack in the recovery due to occlusions is because CIFAR-10 images are small, making it difficult for the networks to recover the information lost in the occlusion.

On the STL-10, in Table \ref{table:pert1} we test our transform augmented networks on five different natural unseen perturbations. We observe that all of our transform augmented networks recover the drop in the performance on unseen perturbations. On rotation-scaling, and snow occlusion, object occlusion augmented convolutions perform the best. On object occlusion Gaussian blur augmented convolutions, for Gaussian noise and Gaussian blur elastic augmented convolutions and for elastic perturbations snow augmented convolutions perform the best.  In contrast with CIFAR-10 on STL-10, our models show significant recovery in the drop for occlusion perturbations too. Hence, our transform augmented convolutions show general robustness on unseen naturally perturbed images. 
\vspace{-3mm}
\subsubsection{Evaluating Robustified Networks on Adversarially Perturbed Images.}
In Table \ref{table:adv1}, we contrast the performance of a standard network with our transform augmented networks. We test the performance for $\approx 10\%$ drop as well as for high drop with the high intensity adversarial perturbations, i.e. $\epsilon=2$.  

On the CIFAR-10, we observe all our transform augmented networks counteract adversarial perturbations causing a $\approx 10\%$ drop, with blur augmented convolutions being the best. While for adversarial perturbations with the $\epsilon=2$ causing a drop of, $\approx 64\%$ only blur augmented convolutions succeed in defending against them. Hence, our proposed robustified networks also help against adversarial perturbations on CIFAR-10.

For STL-10 dataset, our transform augmented network with all the perturbations show resistance against adversarial perturbations. On adversarial perturbations which cause a drop of $\approx 10\%$ elastic augmented, occlusion augmented and blur augmented convolutions show the best resistance. However, for a drop of $\approx57\%$ with, the $\epsilon=2$ elastic augmented convolutions show the best recovery of $\approx 23\%$ with rotation scaling and blur augmented convolutions following it with a recovery of $\approx 13\%$. Thus, our transform augmented networks also defend adversarial perturbations for both low and high drops.

\subsection{Comparison with  Data Augmentation }
To compare our transform augmented convolutions with the data augmentation, we select a perturbation from our global group, i.e. object occlusion and a perturbation from local group, i.e. elastic transform and train a Resnet-152 with  each of them separately. 

\begin{figure}
    \centering
    \includegraphics[width=0.9\linewidth]{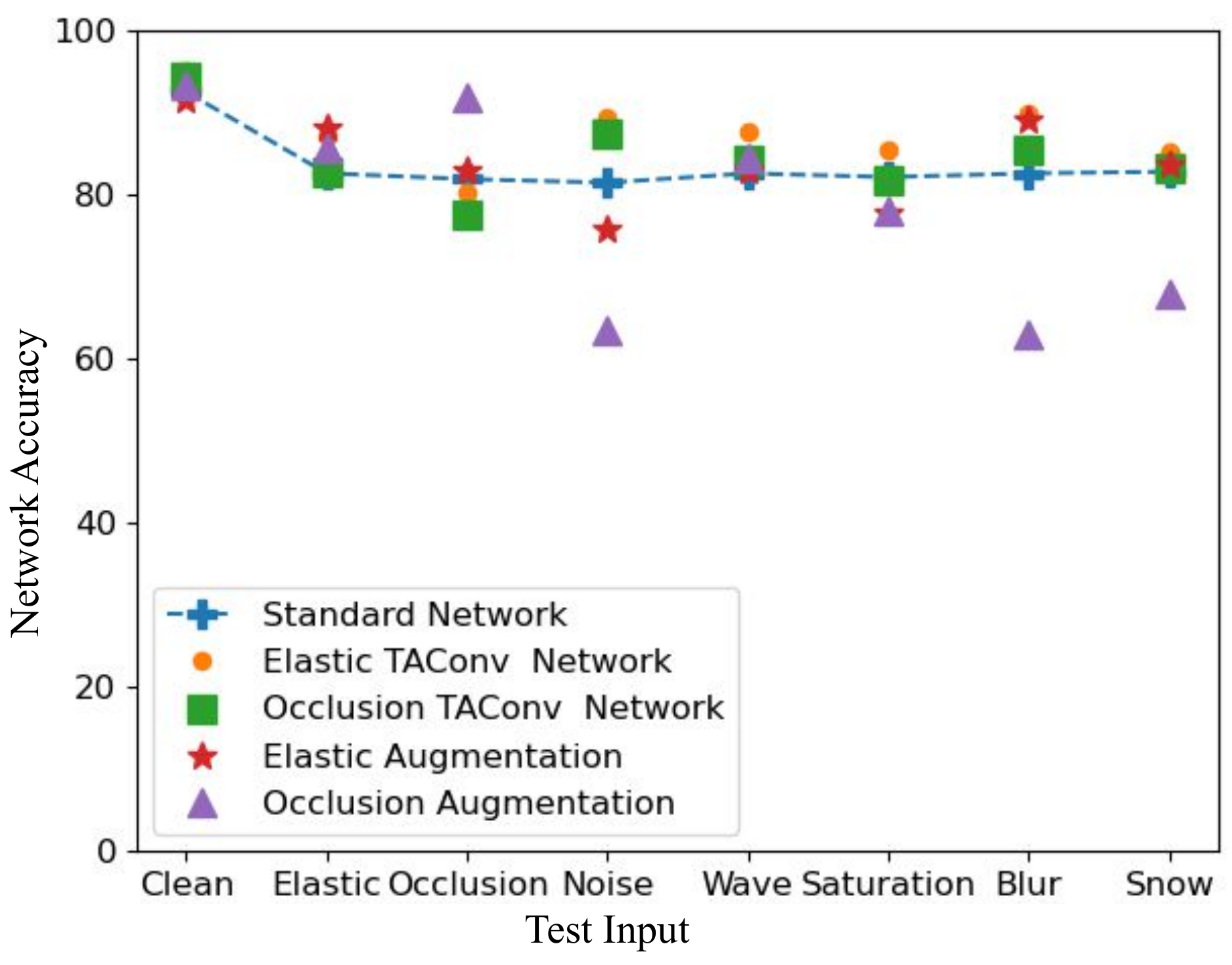}
    \caption{{ Comparing the standard, our TAConv, and data augmented networks on naturally perturbed CIFAR-10.} We observe that although data augmentation shows better recovery for seen perturbations, i.e. occlusion augmentation for occlusion test, but lacks generalization to unseen perturbations. Our TAConv generalizes best to unseen perturbations.}
    \label{fig:Aug}
    \vspace{-2mm}
\end{figure}

Figure \ref{fig:Aug} performs the comparison between a standard, two transform augmented and two data augmented networks on the CIFAR-10. For the clean test, we observe both our TAConv networks show $\approx2\%$ improvement, while with the data augmentation only object occlusion show $\approx0.9\%$ improvement. For naturally perturbed test, although data augmentation shows better recovery for seen perturbations, i.e. occlusion augmentation for occlusion test, but lacks generalization to unseen perturbations  e.g. occlusion augmentation for Gaussian Noise. While, our TAConv generalizes to unseen perturbations.

\vspace{-3mm}
\section{Conclusions}

We formulate a connection between input image perturbations and corresponding transformations of the convolutional filters of a network. We propose a universal approach for integrating any such natural perturbation into a convolutional architecture for enhancing its robustness.

In the evaluation, we first note that our method enhances the network classification accuracy on clean, CIFAR-10 and STL-10 datasets. We conclude that improving the robustness also helps to increase clean accuracy, unexpectedly, where a small loss would also have been acceptable. 

To permit a fair comparison in the performance on perturbed images, we first tune the perturbation parameters to the same drop in classification performance. In this standardized setting, we demonstrate the effectiveness of our method by improving the performance against natural and adversarial perturbations over standard networks. Among the various natural perturbation models, elastic augmented convolutions, corresponding to viewpoint change deformation, generally performs the best. From the experiments, we conclude that TAConv obtains good levels of general robustness.

{\small
\bibliographystyle{ieee_fullname}
\bibliography{egbib}

\begin{thebibliography}{10}\itemsep=-1pt

\bibitem{azulay2018deep}
Aharon Azulay and Yair Weiss.
\newblock Why do deep convolutional networks generalize so poorly to small
  image transformations?
\newblock {\em arXiv preprint arXiv:1805.12177}, 2018.

\bibitem{benz2021revisiting}
Philipp Benz, Chaoning Zhang, Adil Karjauv, and In~So Kweon.
\newblock Revisiting batch normalization for improving corruption robustness.
\newblock In {\em Proceedings of the IEEE/CVF Winter Conference on Applications
  of Computer Vision}, pages 494--503, 2021.

\bibitem{bruna2013invariant}
Joan Bruna and St{\'e}phane Mallat.
\newblock Invariant scattering convolution networks.
\newblock {\em IEEE transactions on pattern analysis and machine intelligence},
  35(8):1872--1886, 2013.

\bibitem{carlini2017towards}
Nicholas Carlini and David Wagner.
\newblock Towards evaluating the robustness of neural networks.
\newblock In {\em 2017 ieee symposium on security and privacy (sp)}, pages
  39--57. IEEE, 2017.

\bibitem{coates2011analysis}
Adam Coates, Andrew Ng, and Honglak Lee.
\newblock An analysis of single-layer networks in unsupervised feature
  learning.
\newblock In {\em Proceedings of the fourteenth international conference on
  artificial intelligence and statistics}, pages 215--223. JMLR Workshop and
  Conference Proceedings, 2011.

\bibitem{cohen2016group}
Taco Cohen and Max Welling.
\newblock Group equivariant convolutional networks.
\newblock In {\em International conference on machine learning}, pages
  2990--2999. PMLR, 2016.

\bibitem{dai2017deformable}
Jifeng Dai, Haozhi Qi, Yuwen Xiong, Yi Li, Guodong Zhang, Han Hu, and Yichen
  Wei.
\newblock Deformable convolutional networks.
\newblock In {\em Proceedings of the IEEE international conference on computer
  vision}, pages 764--773, 2017.

\bibitem{dodge2017quality}
Samuel Dodge and Lina Karam.
\newblock Quality resilient deep neural networks.
\newblock {\em arXiv preprint arXiv:1703.08119}, 2017.

\bibitem{dodge2017study}
Samuel Dodge and Lina Karam.
\newblock A study and comparison of human and deep learning recognition
  performance under visual distortions.
\newblock In {\em 2017 26th international conference on computer communication
  and networks (ICCCN)}, pages 1--7. IEEE, 2017.

\bibitem{dong2020benchmarking}
Yinpeng Dong, Qi-An Fu, Xiao Yang, Tianyu Pang, Hang Su, Zihao Xiao, and Jun
  Zhu.
\newblock Benchmarking adversarial robustness on image classification.
\newblock In {\em Proceedings of the IEEE/CVF Conference on Computer Vision and
  Pattern Recognition}, pages 321--331, 2020.

\bibitem{engstrom2019exploring}
Logan Engstrom, Brandon Tran, Dimitris Tsipras, Ludwig Schmidt, and Aleksander
  Madry.
\newblock Exploring the landscape of spatial robustness.
\newblock In {\em International Conference on Machine Learning}, pages
  1802--1811. PMLR, 2019.

\bibitem{fawzi2015manitest}
Alhussein Fawzi and Pascal Frossard.
\newblock Manitest: Are classifiers really invariant?
\newblock {\em arXiv preprint arXiv:1507.06535}, 2015.

\bibitem{felzenszwalb2009object}
Pedro~F Felzenszwalb, Ross~B Girshick, David McAllester, and Deva Ramanan.
\newblock Object detection with discriminatively trained part-based models.
\newblock {\em IEEE transactions on pattern analysis and machine intelligence},
  32(9):1627--1645, 2009.

\bibitem{geirhos2017comparing}
Robert Geirhos, David~HJ Janssen, Heiko~H Sch{\"u}tt, Jonas Rauber, Matthias
  Bethge, and Felix~A Wichmann.
\newblock Comparing deep neural networks against humans: object recognition
  when the signal gets weaker.
\newblock {\em arXiv preprint arXiv:1706.06969}, 2017.

\bibitem{goodfellow2014explaining}
Ian~J Goodfellow, Jonathon Shlens, and Christian Szegedy.
\newblock Explaining and harnessing adversarial examples.
\newblock {\em arXiv preprint arXiv:1412.6572}, 2014.

\bibitem{gulshad2021natural}
Sadaf Gulshad and Arnold Smeulders.
\newblock Natural perturbed training for general robustness of neural network
  classifiers.
\newblock {\em arXiv preprint arXiv:2103.11372}, 2021.

\bibitem{hendrycks2020many}
Dan Hendrycks, Steven Basart, Norman Mu, Saurav Kadavath, Frank Wang, Evan
  Dorundo, Rahul Desai, Tyler Zhu, Samyak Parajuli, Mike Guo, et~al.
\newblock The many faces of robustness: A critical analysis of
  out-of-distribution generalization.
\newblock {\em arXiv preprint arXiv:2006.16241}, 2020.

\bibitem{hendrycks2019benchmarking}
Dan Hendrycks and Thomas Dietterich.
\newblock Benchmarking neural network robustness to common corruptions and
  perturbations.
\newblock {\em arXiv preprint arXiv:1903.12261}, 2019.

\bibitem{hinton1981parallel}
Geoffrey~F Hinton.
\newblock A parallel computation that assigns canonical object-based frames of
  reference.
\newblock In {\em Proceedings of the 7th international joint conference on
  Artificial intelligence-Volume 2}, pages 683--685, 1981.

\bibitem{jacobsen2016structured}
Jorn-Henrik Jacobsen, Jan Van~Gemert, Zhongyu Lou, and Arnold~WM Smeulders.
\newblock Structured receptive fields in cnns.
\newblock In {\em Proceedings of the IEEE Conference on Computer Vision and
  Pattern Recognition}, pages 2610--2619, 2016.

\bibitem{jaderberg2015spatial}
Max Jaderberg, Karen Simonyan, Andrew Zisserman, and Koray Kavukcuoglu.
\newblock Spatial transformer networks.
\newblock {\em arXiv preprint arXiv:1506.02025}, 2015.

\bibitem{kanazawa2014locally}
Angjoo Kanazawa, Abhishek Sharma, and David Jacobs.
\newblock Locally scale-invariant convolutional neural networks.
\newblock {\em arXiv preprint arXiv:1412.5104}, 2014.

\bibitem{kanbak2018geometric}
Can Kanbak, Seyed-Mohsen Moosavi-Dezfooli, and Pascal Frossard.
\newblock Geometric robustness of deep networks: analysis and improvement.
\newblock In {\em Proceedings of the IEEE Conference on Computer Vision and
  Pattern Recognition}, pages 4441--4449, 2018.

\bibitem{krizhevsky2009learning}
Alex Krizhevsky, Geoffrey Hinton, et~al.
\newblock Learning multiple layers of features from tiny images.
\newblock 2009.

\bibitem{kurakin2016adversarial}
Alexey Kurakin, Ian Goodfellow, and Samy Bengio.
\newblock Adversarial examples in the physical world.
\newblock {\em arXiv preprint arXiv:1607.02533}, 2016.

\bibitem{laptev2016ti}
Dmitry Laptev, Nikolay Savinov, Joachim~M Buhmann, and Marc Pollefeys.
\newblock Ti-pooling: transformation-invariant pooling for feature learning in
  convolutional neural networks.
\newblock In {\em Proceedings of the IEEE conference on computer vision and
  pattern recognition}, pages 289--297, 2016.

\bibitem{laugros2021increasing}
Alfred LAUGROS, Alice Caplier, and Matthieu Ospici.
\newblock Increasing the coverage and balance of robustness benchmarks by using
  non-overlapping corruptions, 2021.

\bibitem{moosavi2016deepfool}
Seyed-Mohsen Moosavi-Dezfooli, Alhussein Fawzi, and Pascal Frossard.
\newblock Deepfool: a simple and accurate method to fool deep neural networks.
\newblock In {\em CVPR}, 2016.

\bibitem{papernot2016limitations}
Nicolas Papernot, Patrick McDaniel, Somesh Jha, Matt Fredrikson, Z~Berkay
  Celik, and Ananthram Swami.
\newblock The limitations of deep learning in adversarial settings.
\newblock In {\em EuroS\&P}. IEEE, 2016.

\bibitem{recht2018cifar}
Benjamin Recht, Rebecca Roelofs, Ludwig Schmidt, and Vaishaal Shankar.
\newblock Do cifar-10 classifiers generalize to cifar-10?
\newblock {\em arXiv preprint arXiv:1806.00451}, 2018.

\bibitem{robey2020model}
Alexander Robey, Hamed Hassani, and George~J Pappas.
\newblock Model-based robust deep learning.
\newblock {\em arXiv preprint arXiv:2005.10247}, 2020.

\bibitem{rusak2020increasing}
Evgenia Rusak, Lukas Schott, Roland Zimmermann, Julian Bitterwolf, Oliver
  Bringmann, Matthias Bethge, and Wieland Brendel.
\newblock Increasing the robustness of dnns against image corruptions by
  playing the game of noise.
\newblock {\em arXiv preprint arXiv:2001.06057}, 2020.

\bibitem{rusak2020simple}
Evgenia Rusak, Lukas Schott, Roland~S Zimmermann, Julian Bitterwolf, Oliver
  Bringmann, Matthias Bethge, and Wieland Brendel.
\newblock A simple way to make neural networks robust against diverse image
  corruptions.
\newblock In {\em European Conference on Computer Vision}, pages 53--69.
  Springer, 2020.

\bibitem{schneider2020improving}
Steffen Schneider, Evgenia Rusak, Luisa Eck, Oliver Bringmann, Wieland Brendel,
  and Matthias Bethge.
\newblock Improving robustness against common corruptions by covariate shift
  adaptation.
\newblock {\em Advances in Neural Information Processing Systems}, 33, 2020.

\bibitem{simard1998transformation}
Patrice~Y Simard, Yann~A LeCun, John~S Denker, and Bernard Victorri.
\newblock Transformation invariance in pattern recognition—tangent distance
  and tangent propagation.
\newblock In {\em Neural networks: tricks of the trade}, pages 239--274.
  Springer, 1998.

\bibitem{sosnovik2021disco}
Ivan Sosnovik, Artem Moskalev, and Arnold Smeulders.
\newblock Disco: accurate discrete scale convolutions.
\newblock {\em arXiv preprint arXiv:2106.02733}, 2021.

\bibitem{sosnovik2021scale}
Ivan Sosnovik, Artem Moskalev, and Arnold~WM Smeulders.
\newblock Scale equivariance improves siamese tracking.
\newblock In {\em Proceedings of the IEEE/CVF Winter Conference on Applications
  of Computer Vision}, pages 2765--2774, 2021.

\bibitem{sosnovik2019scale}
Ivan Sosnovik, Micha{\l} Szmaja, and Arnold Smeulders.
\newblock Scale-equivariant steerable networks.
\newblock {\em arXiv preprint arXiv:1910.11093}, 2019.

\bibitem{su2019one}
Jiawei Su, Danilo~Vasconcellos Vargas, and Kouichi Sakurai.
\newblock One pixel attack for fooling deep neural networks.
\newblock {\em TEVC}, 2019.

\bibitem{szegedy2013intriguing}
Christian Szegedy, Wojciech Zaremba, Ilya Sutskever, Joan Bruna, Dumitru Erhan,
  Ian Goodfellow, and Rob Fergus.
\newblock Intriguing properties of neural networks.
\newblock {\em ICLR}, 2013.

\bibitem{tang2021selfnorm}
Zhiqiang Tang, Yunhe Gao, Yi Zhu, Zhi Zhang, Mu Li, and Dimitris Metaxas.
\newblock Selfnorm and crossnorm for out-of-distribution robustness.
\newblock {\em arXiv preprint arXiv:2102.02811}, 2021.

\bibitem{weiler2018learning}
Maurice Weiler, Fred~A Hamprecht, and Martin Storath.
\newblock Learning steerable filters for rotation equivariant cnns.
\newblock In {\em Proceedings of the IEEE Conference on Computer Vision and
  Pattern Recognition}, pages 849--858, 2018.

\bibitem{wong2020learning}
Eric Wong and J~Zico Kolter.
\newblock Learning perturbation sets for robust machine learning.
\newblock {\em arXiv preprint arXiv:2007.08450}, 2020.

\bibitem{Worrall_2017_CVPR}
Daniel~E. Worrall, Stephan~J. Garbin, Daniyar Turmukhambetov, and Gabriel~J.
  Brostow.
\newblock Harmonic networks: Deep translation and rotation equivariance.
\newblock In {\em Proceedings of the IEEE Conference on Computer Vision and
  Pattern Recognition (CVPR)}, July 2017.

\bibitem{worrall2019deep}
Daniel~E Worrall and Max Welling.
\newblock Deep scale-spaces: Equivariance over scale.
\newblock {\em arXiv preprint arXiv:1905.11697}, 2019.

\bibitem{yin2019fourier}
Dong Yin, Raphael~Gontijo Lopes, Jonathon Shlens, Ekin~D Cubuk, and Justin
  Gilmer.
\newblock A fourier perspective on model robustness in computer vision.
\newblock {\em arXiv preprint arXiv:1906.08988}, 2019.

\bibitem{zhang2019limitations}
Huan Zhang, Hongge Chen, Zhao Song, Duane Boning, Inderjit~S Dhillon, and
  Cho-Jui Hsieh.
\newblock The limitations of adversarial training and the blind-spot attack.
\newblock {\em arXiv preprint arXiv:1901.04684}, 2019.

\end{thebibliography}
}
\newpage
 \begin{figure*}[!h]
    \centering
    \includegraphics[width=\linewidth]{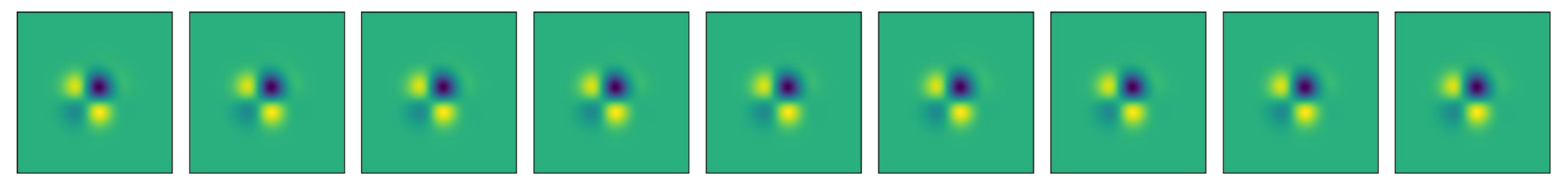}
    \includegraphics[width=\linewidth]{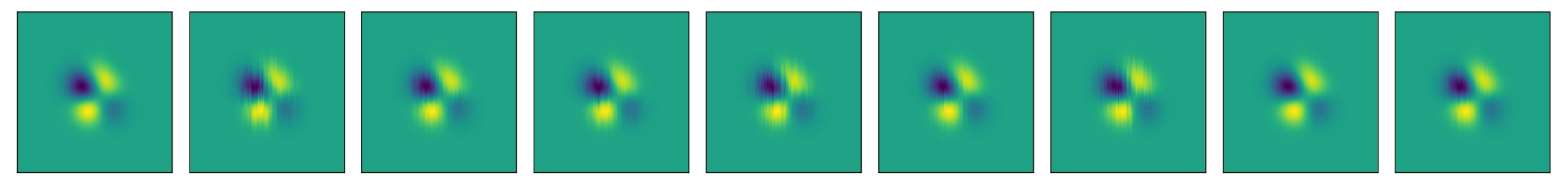}
    \includegraphics[width=\linewidth]{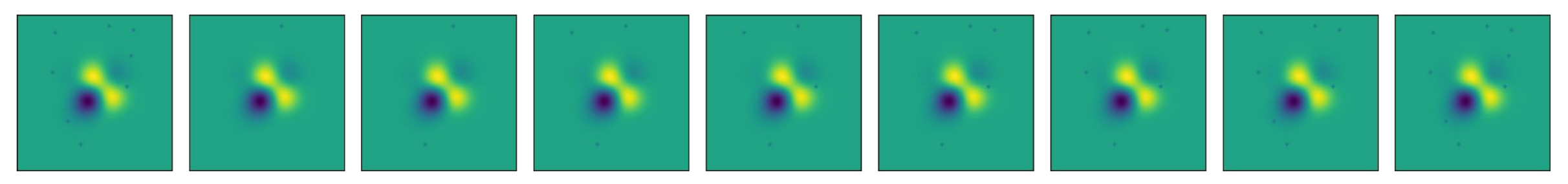}
    \includegraphics[width=\linewidth]{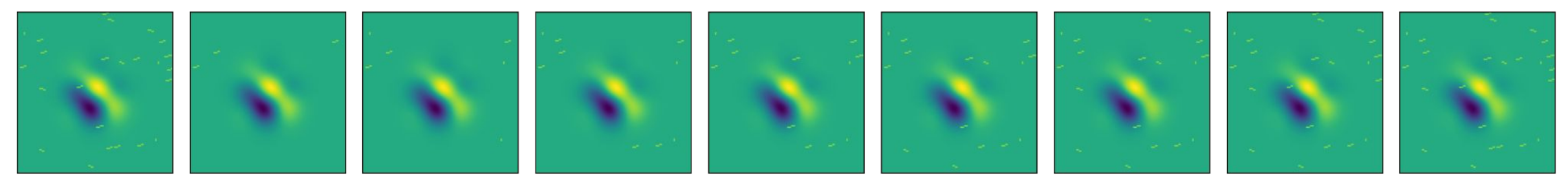}
    \caption{Top: Filter visualizations for varying blur transform. Row 2: Filter visualizations for varying Gaussian Noise transform. 3:  Filter visualizations for varying object occlusion transform. Bottom:  Filter visualizations for varying snow occlusion transform.}
    \label{fig:transform_visualizations}
\end{figure*} 
\section{Appendix}
\subsection{Transformations for Filters}
Here we detail the rest of four transformations used in our experiments to design transform augmented convolutions (TAConv). 

\myparagraph{Gaussian Blur.} We know from the scale-space theory, a smoothed function is the convolution of the original function $f$ with the Gaussian weight function $G$.
\begin{align}
\psi_\sigma(x,y)&=(f\star G_\sigma)(x,y) \\
    G^\sigma(x,y)&=\frac{1}{2\pi\sigma^2}\text{exp}\left[\frac{−x^2+y^2}{2\sigma^2}\right]
\end{align}
The size of the local neighborhood is determined by the scale ($\sigma$) of the Gaussian weight function, see Figure \ref{fig:transform_visualizations} (Top).

\myparagraph{Gaussian Noise.} The Gaussian Noise (see Figure \ref{fig:transform_visualizations} (Row 2)) is implemented as follows:
\begin{align}
    \psi (x',y') &= \psi (x,y) + G^\sigma(x,y) \\
    G^\sigma(x,y)&=\frac{1}{2\pi\sigma^2}\text{exp} \left[\frac{−x^2+y^2}{2\sigma^2}
    \right]
\end{align} 

\myparagraph{Object Occlusion.} 
 We introduce occlusion transforms by creating a circular mask $M_{c_x, c_y}$ with the center being  $c_x$ and $c_y$. The values for $c_x$ and $c_y$ are selected randomly from the discrete uniform distribution $U[low,high)$. The radius of the circular mask is a hyperparameter. All the values in the $M$ are set to zero, see Figure \ref{fig:transform_visualizations} (Row 3). 
 
\myparagraph{Snow Occlusion.} 
 We introduce snow occlusion transforms by creating line shaped $n$ number of masks $M_{x, y, s}$ with $x,y$, $s$ being coordinates of starting and ending point and slope respectively. The number of masks $n$ is a hyperparameter, the slope is selected  from the discrete uniform distribution $U[slope_{low} ,slope_{high} )$, see Figure \ref{fig:transform_visualizations} (Bottom).  

\end{document}


\title{Wiggling Weights for Improving the Robustness of Classifiers}

\author{First Author\\
Institution1\\
Institution1 address\\
{\tt\small firstauthor@i1.org}
\and
Second Author\\
Institution2\\
First line of institution2 address\\
{\tt\small secondauthor@i2.org}
}
\maketitle

 \section{Appendix}
 \begin{figure*}
    \centering
    \includegraphics[width=\linewidth]{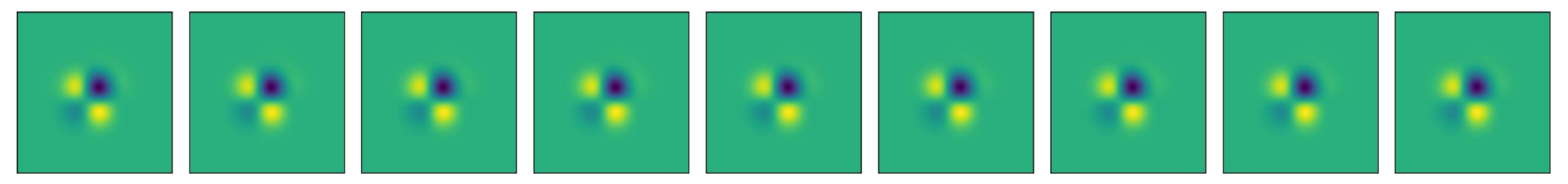}
    \includegraphics[width=\linewidth]{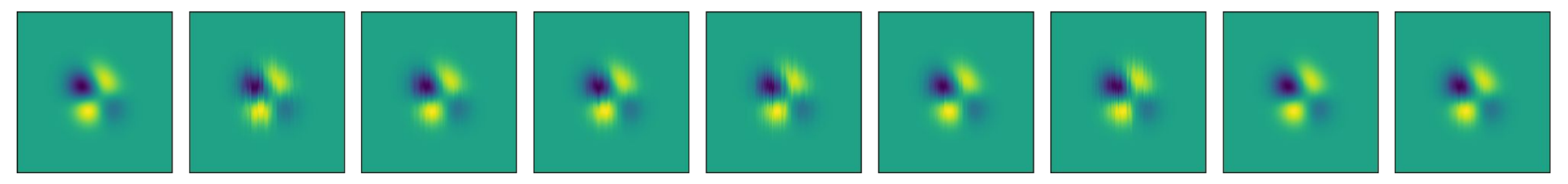}
    \includegraphics[width=\linewidth]{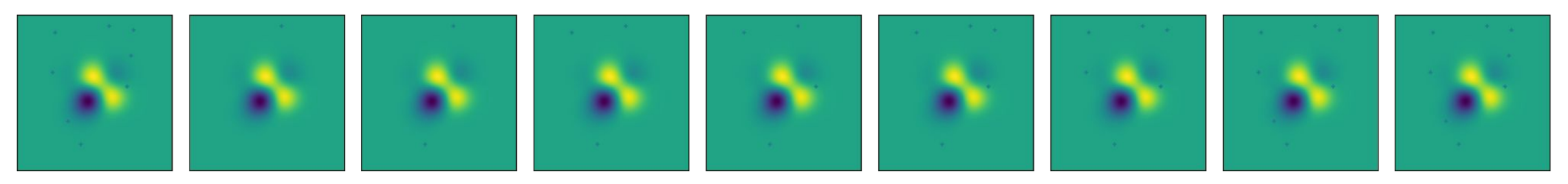}
    \includegraphics[width=\linewidth]{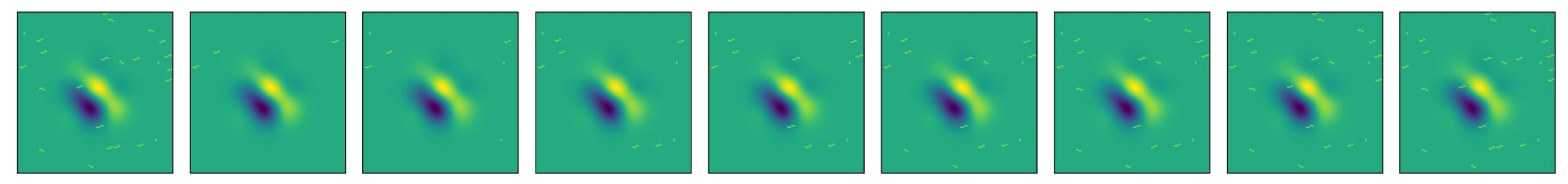}
    \caption{Top: Filter visualizations for varying blur transform. Row 2: Filter visualizations for varying Gaussian Noise transform. 3:  Filter visualizations for varying object occlusion transform. Bottom:  Filter visualizations for varying snow occlusion transform.}
    \label{fig:transform_visualizations}
\end{figure*} 
\subsection{Transformations for Filters}
Here we detail the rest of four transformations used in our experiments to design transform augmented convolutions (TAConv). 

\myparagraph{Gaussian Blur.} We know from the scale-space theory, a smoothed function is the convolution of the original function $f$ with the Gaussian weight function $G$.
\begin{align}
\psi_\sigma(x,y)&=(f\star G_\sigma)(x,y) \\
    G^\sigma(x,y)&=\frac{1}{2\pi\sigma^2}\text{exp}\left[\frac{−x^2+y^2}{2\sigma^2}\right]
\end{align}
The size of the local neighborhood is determined by the scale ($\sigma$) of the Gaussian weight function, see Figure \ref{fig:transform_visualizations} (Top).

\myparagraph{Gaussian Noise.} The Gaussian Noise (see Figure \ref{fig:transform_visualizations} (Row 2)) is implemented as follows:
\begin{align}
    \psi (x',y') &= \psi (x,y) + G^\sigma(x,y) \\
    G^\sigma(x,y)&=\frac{1}{2\pi\sigma^2}\text{exp} \left[\frac{−x^2+y^2}{2\sigma^2}
    \right]
\end{align} 

\myparagraph{Object Occlusion.} 
 We introduce occlusion transforms by creating a circular mask $M_{c_x, c_y}$ with the center being  $c_x$ and $c_y$. The values for $c_x$ and $c_y$ are selected randomly from the discrete uniform distribution $U[low,high)$. The radius of the circular mask is a hyperparameter. All the values in the $M$ are set to zero , see Figure \ref{fig:transform_visualizations} (Row 3). 
 
\myparagraph{Snow Occlusion.} 
 We introduce snow occlusion transforms by creating line shaped $n$ number of masks $M_{x, y, s}$ with $x,y$, $s$ being coordinates of starting and ending point and slope respectively. The number of masks $n$ is a hyperparameter, the slope is selected  from the discrete uniform distribution $U[slope_{low} ,slope_{high} )$, see Figure \ref{fig:transform_visualizations} (Bottom).